\documentclass[letterpaper, 10 pt, conference]{IEEEconf}  

\IEEEoverridecommandlockouts                              

\usepackage{graphics} 
\usepackage{epsfig} 
\usepackage{mathptmx} 
\usepackage{times} 
\usepackage{amssymb}  
\usepackage{siunitx}
\usepackage{subcaption}
\usepackage{newtxtext,newtxmath}
\usepackage{cite}
\makeatletter
\let\NAT@parse\undefined
\makeatother
\usepackage{CJK}
\usepackage{indentfirst}
\usepackage{cases}
\usepackage{hyperref}
\usepackage{booktabs} 
\usepackage{multirow}
\usepackage{xcolor}
\usepackage{xpatch}
\usepackage{ulem}
\usepackage{makecell}
\usepackage{bm}         
\title{\LARGE \bf
Depth Transfer: Learning to See Like a Simulator for Real-World Drone Navigation
}

\author{Hang Yu$^{1}$, Christophe De Wagter$^{1}$ and Guido C. H. E de Croon$^{1}$
\thanks{All authors are with Faculty of Aerospace Engineering, Delft University of Technology, 2629 HS Delft, The Netherlands. (email: {\tt\small h.y.yu@tudelft.nl};  {\tt\small g.c.h.e.deCroon@tudelft.nl}; {\tt\small c.dewagter@tudelft.nl}).}%
}

\begin{document}

\maketitle
\thispagestyle{empty}
\pagestyle{empty}

\begin{abstract}
Sim-to-real transfer is a fundamental challenge in robot reinforcement learning. Discrepancies between simulation and reality can significantly impair policy performance, especially if it receives high-dimensional inputs such as dense depth estimates from vision.
We propose a novel depth transfer method based on domain adaptation to bridge the visual gap between simulated and real-world depth data. A Variational Autoencoder (VAE) is first trained to encode ground-truth depth images from simulation into a latent space, which serves as input to a reinforcement learning (RL) policy. During deployment, the encoder is refined to align stereo depth images with this latent space, enabling direct policy transfer without fine-tuning. We apply our method to the task of autonomous drone navigation through cluttered environments.
Experiments in IsaacGym show that our method nearly doubles the obstacle avoidance success rate when switching from ground-truth to stereo depth input. Furthermore, we demonstrate successful transfer to the photo-realistic simulator AvoidBench using only IsaacGym-generated stereo data, achieving superior performance compared to state-of-the-art baselines.
Real-world evaluations in both indoor and outdoor environments confirm the effectiveness of our approach, enabling robust and generalizable depth-based navigation across diverse domains.
    
\end{abstract}


\section{INTRODUCTION}
 
\begin{figure*}[thpb] 
   \centering
   \begin{minipage}{0.42\textwidth} 
       \subcaptionbox{}{
           \includegraphics[width=\linewidth]{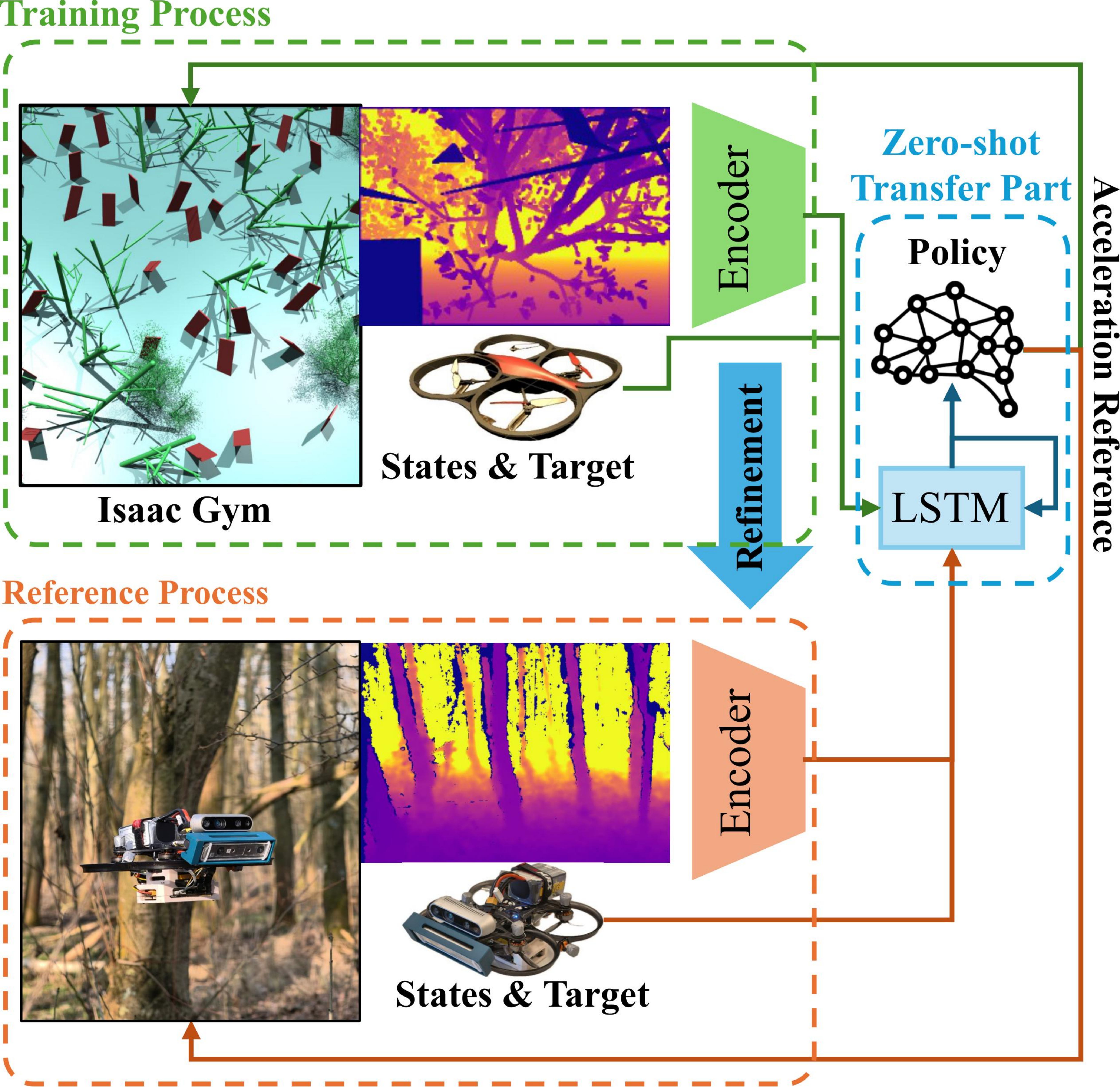}}
   \end{minipage}
   \begin{minipage}{0.52\textwidth} 
       \subcaptionbox{}{
           \includegraphics[width=\linewidth, trim=0 100 0 100, clip]{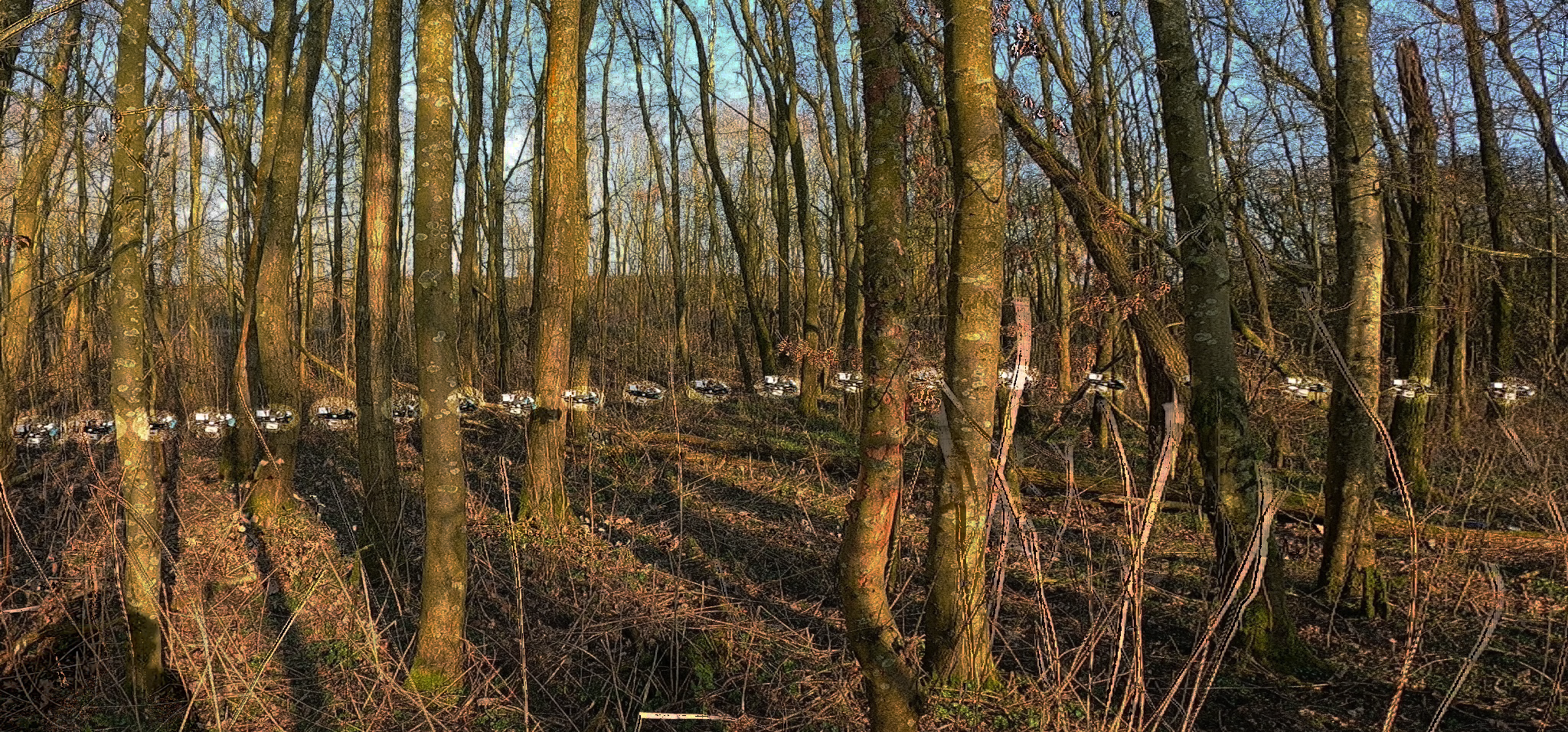}}
           \vspace{5pt} 
       \subcaptionbox{}{
           \includegraphics[width=\linewidth, trim=0 0 0 0, clip]{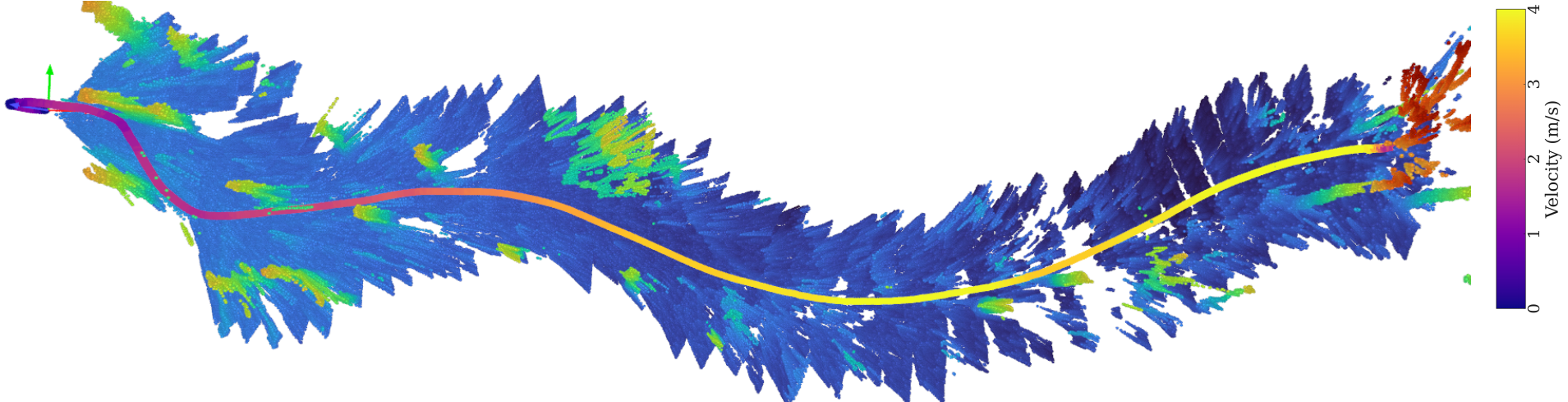}}
   \end{minipage}
   \caption{(a) is the basic framework of the obstacle-free navigation system. When training, the depth images are generated from IsaacGym and the policy is trained in the simulator. During real-world deployment, the depth images are collected from real environments and the policy is evaluated using the refined encoder. (b) illustrates the forest environment for policy evaluation. (c) shows the trajectory from (b), the point clouds here are only for visualization.}
   \label{figure1}
   \vspace{-5mm}
\end{figure*}

Vision-based navigation enables drones to autonomously and safely navigate through complex environments. To reduce the computational burden on onboard systems, reinforcement learning has increasingly been adopted as a key approach for autonomous obstacle avoidance ~\cite{10816139, kaufmann2023champion}. However, real-world training is costly and risky, motivating the use of simulators for safe and efficient policy learning~\cite{song2020flightmare, makoviychuk2021isaac, kulkarni2023aerialgymisaac}.

Among visual modalities, depth maps are particularly favored for sim-to-real transfer, as they offer geometric cues and exhibit better generalization than raw RGB images. Many prior works have successfully trained obstacle avoidance policies using depth maps in simulation and deployed them in challenging real-world environments such as forests~\cite{zhang2024back, 10816139}. However, despite this advantage, a significant perception gap remains between the ideal, noise-free depth maps generated in simulation and the noisy, artifact-laden depth maps captured in the real world. These discrepancies continue to hinder the effectiveness of reinforcement learning when deploying policies trained in simulation to real-world environments.

To address this, we propose a depth transfer method that aligns stereo depth images with simulated ground truth through representation learning and domain adaptation. As shown in Figure~\ref{figure1}(a), we train an encoder to extract latent features from simulated depth images, which are then combined with drone state and navigation target information to train an RL policy.

We evaluated the policy in AvoidBench~\cite{10161097} using an encoder refined with IsaacGym-generated stereo depth. Without using AvoidBench data during training, our method outperforms Ego-Planner~\cite{zhou2020ego} and Agile-Autonomy~\cite{Loquercio2021Science} in all speed settings. It surpasses MAVRL~\cite{10816139} in high-speed scenarios, which was trained in the AvoidBench simulator. For real-world deployment, the encoder is further refined using real stereo depth data, enabling sim-to-real transfer. Figures~\ref{figure1}(b,c) show the evaluation environment and trajectories.

Our main contributions are as follows.
\begin{itemize}
    \item We propose a depth transfer method that aligns stereo depth images with simulator ground truth, enabling RL policies trained entirely in the IsaacGym simulator to generalize effectively to both the photo-realistic simulator AvoidBench and real-world environments. 
    \item We improve latent representation quality by optimizing the VAE architecture and applying min-pooling dilation to better capture obstacle structures in depth maps.
\end{itemize}

\section{Related Work}

\subsection{RL-based Obstacle Avoidance}

Reinforcement learning (RL) has been widely applied to navigation tasks on various robotic platforms, including drones~\cite{kaufmann2023champion, song2023reaching}. Although effective in state-based settings, training RL agents directly from raw pixel data remains challenging due to the high dimensionality of visual inputs. To overcome the difficulty of training vision-based RL policies directly from high-dimensional input, previous work has adopted a two-stage strategy. First, an expert policy is trained using state-based observations, which are typically low-dimensional and easier to learn. Then, a vision-based student policy is trained via imitation learning to mimic the expert~\cite{10160563, xing2024bootstrapping}.

An alternative solution is to use representation learning with staged pretraining, which first extracts compact and informative features from high-dimensional inputs before training the policy, offering improved sample efficiency and generalization compared to fully end-to-end learning. For instance, Mihir \textit{et al.}\cite{10.1007/978-3-031-47966-3_20} introduced the Depth Image-based Collision Encoder (DCE), which addresses the challenge of high-dimensional depth input by compressing it into a compact latent space, while simultaneously enhancing the preservation of fine obstacle details through geometric expansion\cite{kulkarni2023semantically}. 
A well-structured latent space improves both navigation and obstacle avoidance performance. To incorporate temporal context, David \textit{et al.}~\cite{9385894} used an LSTM to predict future depth from historical latent features. Building on this idea, a subsequent approach~\cite{10816139} enhances memory representation by predicting both the previous and current depth images, yielding a more structured and explicit latent space that outperforms memory-free baselines and methods relying solely on future predictions. In this work, we verify that optimizing the VAE architecture to improve depth reconstruction, combined with simple pixel-level dilation, can also preserve more obstacle details and enhance navigation success.

\subsection{Sim-to-Real Transfer and Domain Adaptation}

Multiple techniques have been proposed to improve sim-to-real transfer. To improve generalization, domain randomization has been widely used by introducing physical perturbations in simulators~\cite{kaufmann2023champion, song2023reaching, chebotar2019closing, loquercio2019deep}. While effective for robotic dynamics and obstacle layouts, it struggles with complex visual discrepancies. For instance, \cite{kang2019generalization} proposed using visually diverse simulation data, but this approach suffers from limited coverage and reduced training efficiency. 
Historical Information Bottleneck (HIB)~\cite{he2023bridging} addresses the sim-to-real gap by treating it as an information bottleneck problem and distilling privileged knowledge from historical trajectories to enhance policy generalization.

Stereo vision has a number of challenges that lead to imperfect depth maps, including texture-poor areas, reflections, areas that are occluded in one of the images, etc. This leads to artifacts and invalid regions in stereo depth images, making them substantially different from ground-truth depth images available in simulations. One partial solution to this is to perform stereo vision with images generated in the simulator~\cite{10816139, Loquercio2021Science}. However, doing this during reinforcement learning is computationally expensive and still differs from real-world data.

Domain adaptation provides an alternative by aligning models trained on one domain with data from another. It has been widely explored in computer vision~\cite{hoffman2018cycada, zhou2022domain, peng2022domain}, and more recently in reinforcement learning~\cite{zhu2023transfer}. Many methods adopt a two-stage pipeline: (1) learning domain-invariant representations, followed by (2) RL training~\cite{higgins2017darla, xing2021domain, li2022domain}. However, these methods often require data from multiple domains during training and are typically still validated in visual environments like CarRacing or CARLA~\cite{dosovitskiy2017carla}.

In this work, we apply feature-level domain adaptation to bridge the perception gap between simulated and real-world depth. A VAE is trained on ground-truth depth images in IsaacGym to extract compact latent representations, which are aligned with simulated and real-world stereo depth images via domain adaptation. The RL agent is trained entirely in geometrically simple environments simulation, and evaluated in both the photo-realistic forest environments of AvoidBench~\cite{10161097} and real-world environments using stereo cameras. Experiments demonstrate that our approach significantly improves sim-to-real policy transfer across both settings.

\section{Depth Transfer Based on Domain Adaptation}

Following the training pipeline proposed in~\cite{10816139}, we adopt a similar approach to train an RL agent for obstacle-free navigation. A Variational Autoencoder (VAE) is used to embed high-dimensional depth images into a 64-dimensional latent space, which is then processed by a Long Short-Term Memory (LSTM) module. The LSTM is trained for the reconstruction of both the previous and current depth images, yielding a 256-dimensional latent representation that encodes temporal information. This output is concatenated with the drone's state and target information and used as input to the RL policy.

To bridge the visual gap between simulation and the real world, we propose a depth transfer method based on domain adaptation. This approach aligns the latent spaces of depth images across domains, allowing the policy trained in simulation to be directly deployed in real-world environments.
The full training process consists of the following steps:

\begin{enumerate}
    \item Train an initial PPO policy with few obstacles in IsaacGym using ground-truth depth images for a short time, with randomly initialized VAE and LSTM.
    \item Collect depth images with the initial policy and train the VAE to map into a 64-dimensional latent space.
    \item Train the LSTM to reconstruct both past and current depth images from the latent embeddings.
    \item Retrain the PPO policy using the LSTM output concatenated with the drone's state and target position.
    \item Refine the VAE using stereo depth images from IsaacGym, AvoidBench, or real world to align their latent space with that of the simulated ground truth.
    \item Evaluate the policy in AvoidBench and real-world environments using the refined encoder.
\end{enumerate}

\subsection{Latent Representation Learning}

\begin{figure}[thpb]
    \vspace{-3mm}
    \centering
    {\includegraphics[width=3.2in]{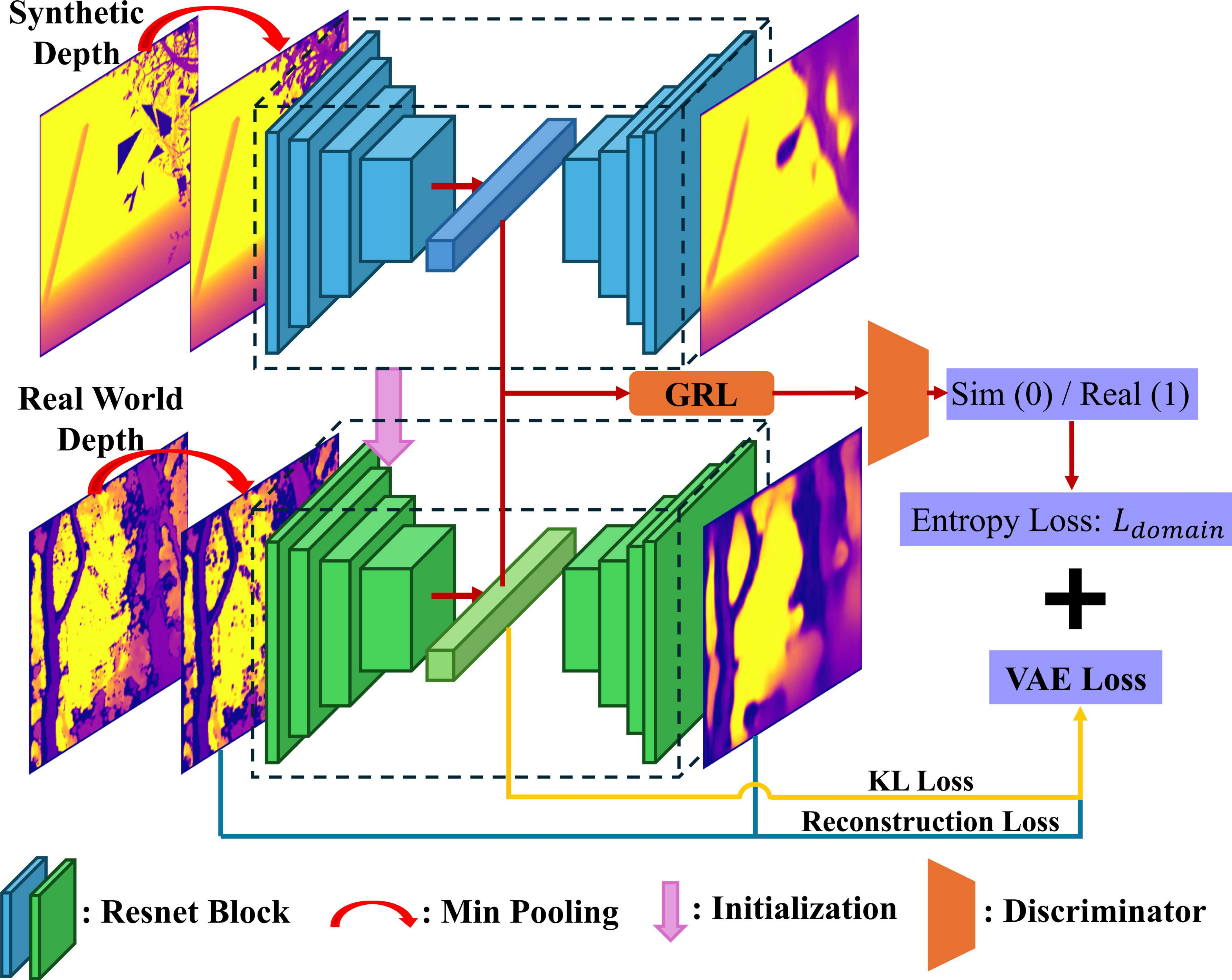}}
    \caption{Domain adaptation for depth transfer. The VAE is trained using depth images from IsaacGym. The encoder maps depth images to a latent space, while the decoder reconstructs the depth images. The GRL and discriminator are used to align the latent spaces of simulated and real-world depth images.}
    \label{fig: network}
    \vspace{-3mm}
\end{figure}

Firstly, we employ the VAE to learn a latent representation of depth images. The VAE consists of an encoder \(E\) and a decoder \(D\). The encoder maps depth images \(X\) to a latent space \(\mathbf{z}_\text{vae}\), while the decoder reconstructs the depth images from this latent representation. The VAE is trained by minimizing both the reconstruction loss and the Kullback-Leibler (KL) divergence between the learned latent distribution and a prior distribution. 

Unlike previous work \cite{10816139}, which relied solely on convolutional layers for feature extraction, we incorporate a residual neural network (ResNet) architecture to enhance the quality of the learned latent space. The ResNet architecture consists of a series of residual blocks, each containing two convolutional layers with skip connections. These skip connections facilitate gradient flow, allowing the network to learn more complex features and improving the reconstruction quality of depth images. In addition, to solve the problem of thin obstacles in depth maps, we apply a simple pixel-level dilation to expand the obstacle sizes in the depth images, which we call the min-pooling dilation.
The min-pooling dilation can be expressed using max-pooling as follows:
\vspace{-2mm}
\begin{equation}
    Y_{i,j} = \min_{(m,n) \in \mathcal{R}_{i,j}} X_{m,n}
\end{equation}
where \( X \) is the input depth map, \( \mathcal{R}_{i,j} \) represents the receptive field (i.e., the pooling window) centered at position \( (i,j) \), and \( Y \) is the resulting output after the min-pooling operation.

As shown in the top part of Figure~\ref{fig: network}, the VAE is trained on depth images from IsaacGym to extract compact latent representations. The encoder consists of a 16-channel convolutional layer followed by six residual blocks, and outputs the mean and variance of the latent distribution. The decoder mirrors the encoder to reconstruct the input depth image. The VAE is optimized with a combination of reconstruction and KL divergence losses.


As illustrated in Figure~\ref{figure1}(a), the latent vector from the encoder is used as input to the LSTM. Following the architecture in~\cite{10816139}, the LSTM is trained to reconstruct both the previous and current depth images to capture temporal context. Its output is concatenated with the drone's state and target position to train the PPO policy. The LSTM is optimized using the following loss:
\begin{equation}
    \mathcal{L}_{\text{LSTM}} = \sum_{i=-1,0} \text{MSE}(Y_{t+iT}, \hat{Y}_{t+iT}), 
    \label{eq2}
\end{equation}
where \(\hat{Y}_{t+iT}\) denotes the reconstructed depth image at time \(t+iT\) from the latent input \(\mathbf{z}_t\), and \(T\) is the temporal interval between prediction steps. Following the setting validated in~\cite{10816139}, we set \(T = 20\).

\subsection{Sim-to-Real Depth Transfer}

After training the policy in IsaacGym, we collect stereo depth images from simulators or real-world environments and refine the VAE to align the resulting latent space with that learned from ground-truth depth in simulation, as shown in the bottom part of Figure~\ref{fig: network}. Before encoding, stereo depth images are preprocessed with min-pooling dilation to enlarge obstacle regions. The VAE is initialized with weights from simulation and fine-tuned using the collected stereo data.

To bridge the domain gap, we adopt a Gradient Reversal Layer (GRL) and a discriminator. The GRL is placed between the latent space and discriminator to reverse gradients during backpropagation, encouraging the encoder to generate domain-invariant representations. The discriminator is trained to distinguish between simulated and real-world latent features, thus guiding the encoder toward alignment. 
During the forward pass, the GRL behaves as an identity function, i.e., \( \mathbf{z}_{\text{grl}} = \mathbf{z}_{\text{vae}} \).
where \( \mathbf{z}_{\text{grl}} \) is the output of the GRL.
During backpropagation, the gradient of the loss function \( L \) with respect to the input is reversed and scaled by \( \lambda_{\text{grl}} \):
\begin{equation}
\frac{\partial L}{\partial \mathbf{z}_{\text{vae}}} = -\lambda_{\text{grl}} \frac{\partial L}{\partial \mathbf{z}_{\text{grl}}}
\end{equation}
where \( \lambda_{\text{grl}} \) is a hyperparameter that controls the strength of the gradient reversal.

The discriminator is a binary classifier consisting of two fully connected layers followed by a log-softmax layer. It outputs a probability \( p \) indicating whether the input latent vector originates from the ground depth or the stereo depth. The entropy loss of the discriminator is combined with the VAE loss to update the VAE for stereo depth images. The loss function for the VAE is then updated as follows:
\begin{equation}
    \begin{aligned}
    \mathcal{L}_{\text{VAE}} = \mathcal{L}_{\text{recon}} + \beta \mathcal{L}_{\text{KL}} + \gamma \mathcal{L}_{\text{domain}}
    \end{aligned}
    \label{eq3}
 \end{equation}
where \( \mathcal{L}_{\text{domain}} \) is the domain loss, and \( \gamma \) is a hyperparameter that balances the domain loss with the reconstruction loss and KL divergence. 
 
The domain adaptation loss \( \mathcal{L}_{\text{domain}} \) is defined as a binary cross-entropy loss applied to the discriminator, which distinguishes whether a latent vector \( \mathbf{z}_\text{vae} \) comes from ground-truth or stereo depth images:
\begin{equation}
    \mathcal{L}_{\text{domain}} = - \mathbb{E}_{\mathbf{z}_\text{vae}^{\text{gt}}} \log D_\text{dis}(\mathbf{z}_\text{vae}^{\text{gt}}) 
    - \mathbb{E}_{\mathbf{z}_\text{vae}^{\text{stereo}}} \log \big(1 - D_\text{dis}(\mathbf{z}_\text{vae}^{\text{stereo}})\big),
\end{equation}
where \( \mathbf{z}_\text{vae}^{\text{gt}} = E(X^{\text{gt}}) \) and \( \mathbf{z}_\text{vae}^{\text{stereo}} = E(X^{\text{stereo}}) \) are latent vectors extracted from ground-truth and stereo depth images, respectively. The discriminator \( D_\text{dis}(\cdot) \) outputs the probability that the input belongs to the ground-truth domain, and \( \mathbb{E}[\cdot] \) denotes the batch-wise expectation.

Since the GRL inverts the gradient during backpropagation, the encoder is trained adversarially to fool the discriminator, encouraging the latent representations of ground truth and stereo depth images to become indistinguishable.

\section{Reinforcement Learning for Obstacle Avoidance}

In this section, we describe the RL training process for obstacle avoidance. The RL agent is trained using the latent representation learned from the VAE and LSTM, combined with the drone's state and target information. The agent is trained using the PPO algorithm, which maximizes the expected cumulative reward by updating the policy parameters.

\subsection{Reference Generator} 

We use the Aerial Gym simulator \cite{kulkarni2023aerialgymisaac} for training the RL agent. Aerial Gym is built on top of the IsaacGym physics engine, which provides a drone model with realistic dynamics. The drone model is described as a rigid body with mass \( m \), and a set of propellers that generate thrust.
The motion of the drone is governed by the following rigid body dynamics:
\begin{equation}
    \label{eq:rigid_body_dyn}
        \begin{bmatrix}
            \dot{\bm{p}}_{\mathcal{W}} \\
            \dot{\bm{q}}_{\mathcal{WB}} \\
            \dot{\bm{v}}_{\mathcal{W}} \\
            \dot{\bm{\omega}}_{\mathcal{B}}
        \end{bmatrix} = \begin{bmatrix}
            \bm{v}_{\mathcal{W}} \\
            \frac{1}{2}\bm{q}_{\mathcal{WB}} \otimes \begin{bmatrix}
                0 & \bm{\omega}_{\mathcal{B}}^{\mathsf{T}}
            \end{bmatrix}^{\mathsf{T}} \\
            \bm{g}_{\mathcal{W}} + \frac{1}{m} \bm{R}_{\mathcal{WB}} (\bm{f}_{a} + \bm{f}_{d}) \\
            \bm{J}^{-1} (\bm{\tau}_{a} + \bm{\tau}_{d} - \bm{\omega}_{\mathcal{B}} \times \bm{J} \bm{\omega}_{\mathcal{B}})
        \end{bmatrix},
\end{equation}
where \( \bm{p}_{\mathcal{W}} \), \( \bm{q}_{\mathcal{WB}} \), \( \bm{v}_{\mathcal{W}} \), and \( \bm{\omega}_{\mathcal{B}} \) represent the drone's position, orientation (quaternion), linear velocity, and angular velocity, respectively. The thrust and torque produced by the propellers are denoted by \( (\bm{f}_{a} \), \( \bm{\tau}_{a}) \), while \((\bm{f}_d, \bm{\tau}_d)\) represent the aerodynamic drag. \( \bm{g}_{\mathcal{W}} \) is the gravitational acceleration in the world frame, while \( \bm{J} \) is the drone’s inertia matrix. The rotation matrix \( \bm{R}_{\mathcal{WB}} \) transforms vectors from the body frame to the world frame.

We use the acceleration \(\bm{a}_{\text{cmd}}\) and yaw rate \(\dot{\psi}\) as the action space for the RL agent. A reference generator maps these commands into smooth attitude and thrust references to ensure feasible tracking. The desired roll (\(\phi_{\text{sp}}\)), pitch (\(\theta_{\text{sp}}\)), and yaw (\(\psi_{\text{sp}}\)) angles are computed as:
\vspace{-2mm}
\begin{equation}
    \label{eq:ref_gen}
    \begin{aligned}
        \theta_{\text{sp}} &= \tan^{-1} \left( \frac{a_{x,\text{cmd}}}{a_{z,\text{cmd}}} \right), \quad
        \phi_{\text{sp}} = \tan^{-1} \left( \frac{-a_{y,\text{cmd}}}{\sqrt{a_{z,\text{cmd}}^2 + a_{x,\text{cmd}}^2}} \right), \\
        \psi_{\text{sp}} &= \psi + \dot{\psi} \Delta t,
    \end{aligned}
\end{equation}

\noindent
where \( a_{x,\text{cmd}}, a_{y,\text{cmd}}, a_{z,\text{cmd}} \) are body-frame accelerations, and \( \Delta t \) (\(0.01\,\text{s}\)) is the simulation time step.

To generate a smooth angular velocity reference, we define the attitude error \(\bm{e}_\theta\) based on the desired quaternion \( \bm{q}_{\mathcal{WB},\text{sp}} \), and update the reference angular velocity using a second-order underdamped system with saturation constraints:
\vspace{-2mm}
\begin{equation}
    \label{eq:ref_update}
    \begin{aligned}
        \bm{e}_\theta &= \bm{q}_{\mathcal{WB}} - \bm{q}_{\mathcal{WB},\text{sp}}, \\
        \dot{\bm{\omega}}_{\mathcal{B}, \text{sp}} &= -2 \bm{\omega}_{\text{n}} \bm{\zeta} \bm{\omega}_{\mathcal{B}, \text{sp}} - \bm{\omega}_{\text{n}}^2 \bm{e}_\theta, \\
        \dot{\bm{\omega}}_{\mathcal{B}, \text{sp}} &= \text{clip}(\dot{\bm{\omega}}_{\mathcal{B}, \text{sp}}, -\dot{\bm{\omega}}_{\max}, \dot{\bm{\omega}}_{\max}), \\
        \bm{\omega}_{\mathcal{B}, \text{sp}} &= \bm{\omega}_{\mathcal{B}, \text{sp}} + \dot{\bm{\omega}}_{\mathcal{B}, \text{sp}} \Delta t, \\
        \bm{\omega}_{\mathcal{B}, \text{sp}} &= \text{clip}(\bm{\omega}_{\mathcal{B}, \text{sp}}, -\bm{\omega}_{\max}, \bm{\omega}_{\max}).
    \end{aligned}
\end{equation}
Here, \( \bm{\omega}_{\text{n}} \) is the natural frequency controlling convergence speed, and \( \bm{\zeta} \) is the damping ratio that balances responsiveness and overshoot. \( \dot{\bm{\omega}}_{\max} \) and \( \bm{\omega}_{\max} \) limit angular acceleration and velocity to ensure physical feasibility and stability.

Similarly, the thrust is updated using a rate-limited approach:
\vspace{-3mm}
\begin{equation}
    \dot{T}_{\text{sp}} = \text{clip} \left( \frac{T_{\text{sp}} - T}{T_{\text{sim}}}, -\dot{T}_{\max}, \dot{T}_{\max} \right), \quad
    T_{\text{sp}} \leftarrow T + \dot{T}_{\text{sp}} \Delta t.
\end{equation}

\noindent
where \( T_{\text{sim}} \) is the RL action duration (\(\text{0.1}s\) in our case), and \( \dot{T}_{\max} \) is the maximum allowable thrust rate.

The reference linear acceleration, linear velocity, position, orientation, and angular velocity are then updated through Equation \ref{eq:rigid_body_dyn}.
This ensures that the reference states evolve smoothly, aligning with the drone’s rigid body dynamics while maintaining stability and responsiveness.

\subsection{Task Formulation}

The obstacle avoidance task is formulated as a Markov Decision Process (MDP), defined by the tuple \( \langle \mathcal{S}, \mathcal{A}, \mathcal{P}, \mathcal{R}, \gamma \rangle \).  
Here, \( \mathcal{S} \) denotes the state space, which includes the drone’s state, target information, and the latent representation extracted from depth images. The action space \( \mathcal{A} \) is defined as \( \bm{a} = [\bm{a}_{\text{cmd}}, \dot{\psi}] \), where \( \bm{a}_{\text{cmd}} \) is the commanded acceleration and \( \dot{\psi} \) is the yaw rate.

\subsubsection{Transition Probability}

In our case, the transition function \( \mathcal{P} \) is deterministic. Given an action command \( \bm{a} \), the reference generator (Equations~\ref{eq:ref_gen}, \ref{eq:ref_update}, and \ref{eq:rigid_body_dyn}) will generate a smooth reference trajectory. The low-level controller maps the action commands to the desired thrust and torque, which are sent to the drone model. The drone's states are then updated using rigid body dynamics simulated by the physics engine.

\subsubsection{Observations}
At each time step \( t \), the observation \( \bm{o}_t \) consists of a 14-dimensional physical state vector and a 256-dimensional perception vector obtained from the VAE and LSTM. The full observation is defined as:
\begin{equation}
    \vspace{-2mm}
    \bm{o}_t = \left[
        \log d_{\text{hor}},\ 
        d_z,\ 
        \bm{d}_{\text{norm}},\ 
        \bm{v}_{\mathcal{W}},\ 
        \bm{\omega}_{\mathcal{B}},\ 
        \dot{\bm{\omega}}_{\mathcal{B}},\ 
        \bm{z}_t
    \right],
\end{equation}
where \( d_{\text{hor}} \) and \( d_z \) denote the horizontal and vertical distances to the target; \( \bm{d}_{\text{norm}} \) is the normalized direction vector to the target; \( \bm{v}_{\mathcal{W}} \) is the linear velocity in the world frame; \( \bm{\omega}_{\mathcal{B}} \) and \( \dot{\bm{\omega}}_{\mathcal{B}} \) are the angular velocity and angular acceleration in the body frame; and \( \bm{z}_t \) is the 256-dimensional latent representation of the depth image.

\subsubsection{Reward Function}

The reward function consists of a \textbf{progress reward} \( r_{\text{prog}} \) and several \textbf{terminal rewards}. The progress reward is defined as:
\begin{equation}
    \label{eq:progress_reward}
    \begin{aligned}
    r_{\text{prog}} = &\ \lambda_{d} \cdot \log d_{\text{hor}} + \lambda_{z} \cdot d_{z} + \max(0,\ v_{\text{hor}} - v_{\text{max}}) \cdot \lambda_v \cdot v_{\text{hor}} \\
    & + \lambda_{\text{dir}} \cdot \|\bm{d}_{\text{norm}} - \bm{v}_{\text{norm}}\|_1 + \lambda_{\text{input}} \cdot \|\bm{\omega}_{\mathcal{B}}\| \\
    & + \lambda_{\text{perc}} \cdot \left( |v_{\mathcal{B}, y}| + \max(0,\ -v_{\mathcal{B}, x}) \right),
    \end{aligned}
\end{equation}

\noindent
where \( \lambda_d \), \( \lambda_z \), \( \lambda_v \), \( \lambda_{\text{dir}} \), \( \lambda_{\text{input}} \), and \( \lambda_{\text{perc}} \) are negative hyperparameters that shape the agent's behavior.

The first two terms penalize the horizontal and vertical distances to the target, encouraging the drone to approach it.  
The third term penalizes excessive horizontal velocity \( v_{\text{hor}} \) when it exceeds the predefined limit \( v_{\text{max}} \), ensuring safe and controlled motion.  
The fourth term penalizes the angular deviation between the normalized distance vector \( \bm{d}_{\text{norm}} \) and velocity vector \( \bm{v}_{\text{norm}} \), guiding the drone to move directly toward the target.  
The fifth term penalizes high angular velocity \( \bm{\omega}_{\mathcal{B}} \), promoting stable attitude control.  
Finally, the last term introduces a perception-aware penalty by discouraging lateral and backward motion in the body-frame velocity \( \bm{v}_{\mathcal{B}} = [v_{\mathcal{B}, x},\ v_{\mathcal{B}, y},\ v_{\mathcal{B}, z}] \). This encourages the drone to maintain a forward-facing orientation, thereby improving perception and enhancing obstacle avoidance performance.

This progress reward is designed to balance goal-reaching efficiency, flight stability, and safety, guiding the drone toward the target while maintaining a favorable attitude.

The overall reward function, composed of the progress reward and several terminal rewards, is defined as:
\begin{equation}
    \label{eq:reward_functions}
    r = 
    \begin{cases}
        r_{\text{exceed}},     & \text{if } (p_t < p_{\text{min}} \text{ or } p_t > p_{\text{max}}),\quad p \in \{x, y, z\} \\
        r_{\text{arrive}},     & \text{if } \|\bm{d}\| < d_{\text{min}} \\
        r_{\text{collision}},  & \text{if collision occurs} \\
        r_{\text{prog}},       & \text{otherwise}
    \end{cases}
\end{equation}

\noindent
where \( p_{\text{min}} \) and \( p_{\text{max}} \) define the allowable position boundaries, and \( r_{\text{exceed}} \) is the penalty for flying out of bounds.  
The term \( d_{\text{min}} \) is the arrival threshold, and \( r_{\text{arrive}} \) denotes the reward for successfully reaching the target.  
The term \( r_{\text{collision}} \) penalizes collisions with obstacles, while \( r_{\text{prog}} \) refers to the progress reward defined in Equation~\ref{eq:progress_reward}.

\section{Experiments}

\subsection{Depth Reconstruction with Min-Pooling Dilation}

\begin{figure}[thpb]
    \centering
    {\includegraphics[width=3.4in]{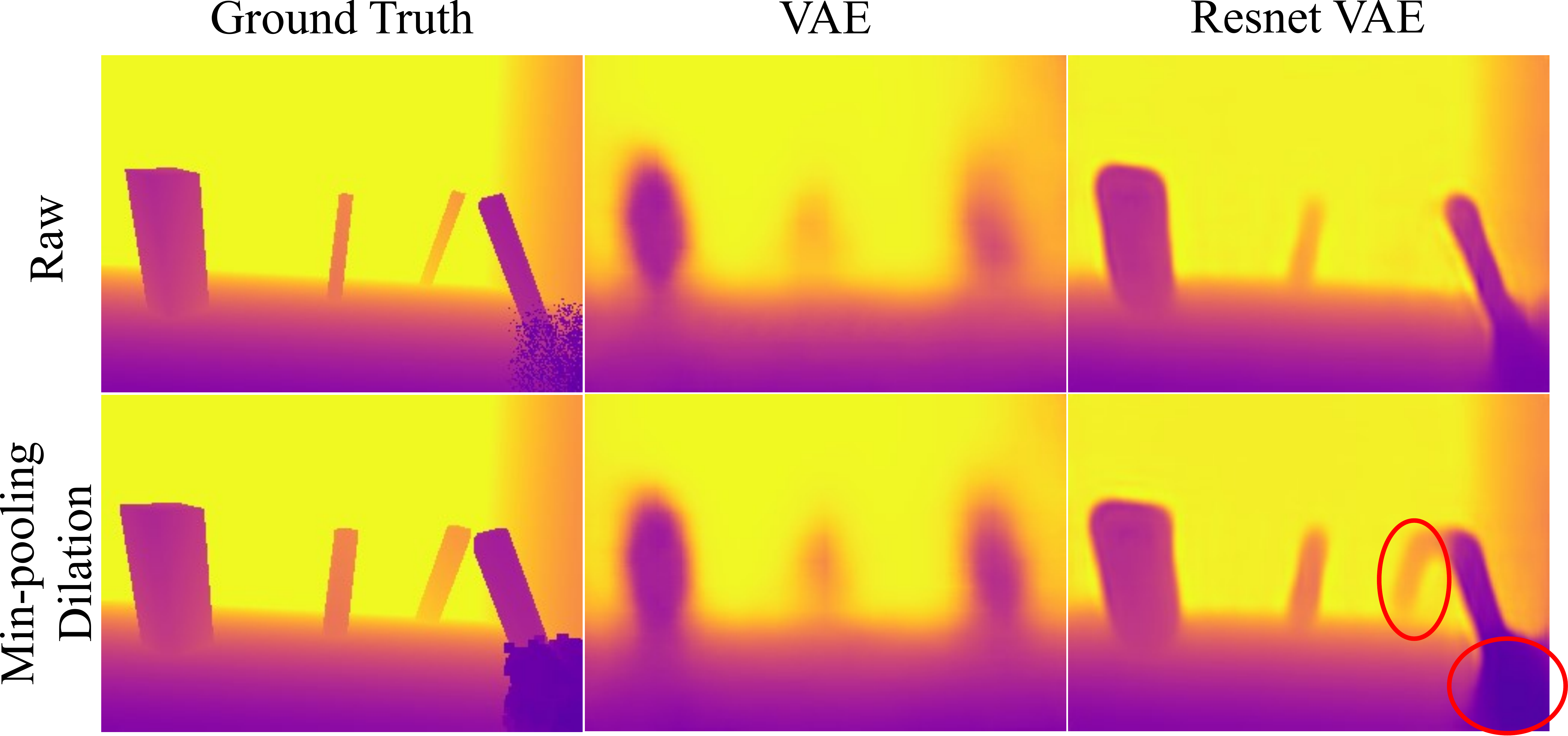}}
    \caption{Depth reconstruction with min-pooling dilation. The first column shows the original depth and after dilation. The second and third columns compare the reconstructed depth images from a standard VAE and a VAE with a ResNet architecture.}
    \label{fig: reconstruction}
    \vspace{-3mm}
\end{figure}

Different from prior work~\cite{10816139}, we apply min-pooling dilation to expand obstacle regions in the depth images. As illustrated in the first column of Figure~\ref{fig: reconstruction}, this operation effectively enlarges obstacles, making them more prominent and easier to detect. The second and third columns compare the reconstructed depth images from a standard VAE (as used in~\cite{10816139}) and a VAE with a ResNet architecture. The ResNet VAE significantly improves reconstruction fidelity, producing outputs that are closer to the ground truth. Moreover, when combined with min-pooling dilation, the reconstruction quality further improves—particularly in areas close to the drone—by preserving fine-grained details and better separating obstacles from the background.

To evaluate the effectiveness of the ResNet-based VAE and min-pooling dilation in obstacle avoidance, we trained four policies using different latent representations: standard VAE, VAE with dilation, ResNet VAE, and ResNet VAE with dilation. The evaluation was conducted on 24 maps of varying complexity, each with 10 trials. As shown in Table~\ref{tab:dilation_results}, both the ResNet architecture and min-pooling significantly improve performance, with their combination achieving the highest success rate. Although the absolute success rates may appear modest, the obstacle avoidance task in our evaluation is highly challenging, with dense and randomized obstacle layouts.

\begin{table}[t]
    \small 
    \centering
    \caption{Ablation studies for Resnet VAE and min-pooling dilation}
    \label{tab:dilation_results}
    \begin{tabular}{l c c c c}
        \toprule
        & VAE & \makecell{VAE \& \\ Dilation} & Resnet VAE & \makecell{Resnet VAE \& \\ Dilation} \\
        \midrule
        \makecell{Success \\ Rate (\%)} & 50.4 & 64.2 & 59.6 & \textbf{71.7} \\
        \bottomrule
    \end{tabular}
    \vspace{-4mm}
\end{table}

\subsection{Visualization and Analysis of Latent Representations}

The RL policy will be trained in IsaacGym with ground truth depth images, since using the stereo depth means the simulator always need to render two images and run the stereo matching algorithm, which is time-consuming and not efficient. The goal of the domain adaptation is to align the latent spaces of the depth images from different sources, so that the policy trained in IsaacGym can be directly transferred to the visually more realistic AvoidBench and real-world environments.

\begin{figure}[thpb]
    \vspace{-2mm}
    \centering
    {\includegraphics[width=3.4in, trim=10 10 10 10, clip]{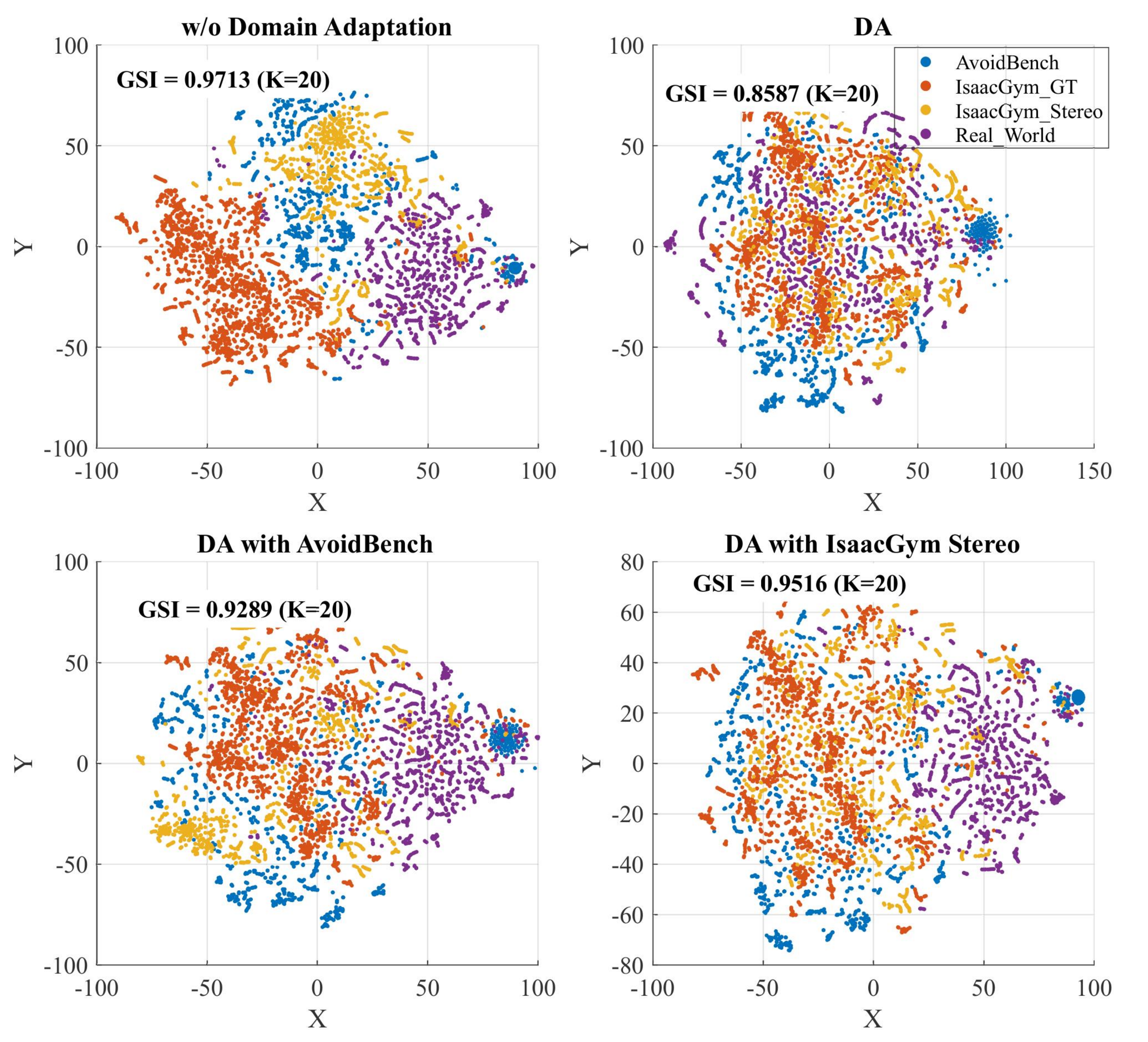}}
    \caption{t-SNE visualization of latent space alignment. The plot includes ground truth depth images from IsaacGym (orange), stereo depth images from IsaacGym (yellow), stereo depth images from AvoidBench (blue), and real-world depth images (purple).}
    \label{fig: tSNE}
    \vspace{-3mm}
\end{figure}

To evaluate the effectiveness of feature-level domain adaptation, we visualize the latent space using t-SNE~\cite{van2008visualizing}, as shown in Figure~\ref{fig: tSNE}. The plot includes four categories: ground truth depth from IsaacGym (orange), stereo depth from IsaacGym (yellow), AvoidBench (blue), and real-world environments (purple).

In the top-left plot, all categories share the same encoder trained from ground truth depth, resulting in clear domain separation—especially between ground truth and three stereo-style dept categories. After applying domain adaptation to each stereo-style depth category (top-right), the distributions become more aligned, demonstrating effective feature-level adaptation.

The bottom plots examine whether an encoder refined with IsaacGym or AvoidBench stereo depth can generalize to real-world inputs. Compared to the top-right configuration (with dedicated encoders), both single-encoder variants show poorer alignment with real-world depth images.

We quantify the alignment using the Geometric Separability Index (GSI)~\cite{greene2001feature}, where lower values indicate better inter-class mixing. We computed the GSI values for the ground truth depth images from IsaacGym and real-world depth images under four different encoder configurations, as reported in each plot of Figure \ref{fig: tSNE}. The configuration where all categories use different encoders achieves the lowest GSI value (0.8587), indicating the best latent space alignment. In contrast, the configuration where the encoder was refined using stereo depth images from AvoidBench results in a higher GSI value (0.9289), though still lower than the baseline where all categories use the same encoder (0.9713). Refining the encoder with stereo depth images from IsaacGym also improves alignment (0.9516) for real-world depth images, though it is less effective than using AvoidBench for adaptation.

These results confirm that domain adaptation effectively aligns the latent spaces of different depth inputs, with the best alignment achieved using separate encoders or AvoidBench-refined features for real-world deployment.

\subsection{Zero-shot Policy Transfer in IsaacGym}

We assess the effectiveness of our depth transfer method by deploying a policy trained on ground truth depth images to stereo depth images in IsaacGym. Evaluation is conducted across eight maps with different obstacle layouts, each comprising ten trials with randomized start and target positions. The success rate is defined as the percentage of trials in which the drone reaches the target without collisions.

\begin{figure*}[tpb]
    \centering
    {\includegraphics[width=6.6in]{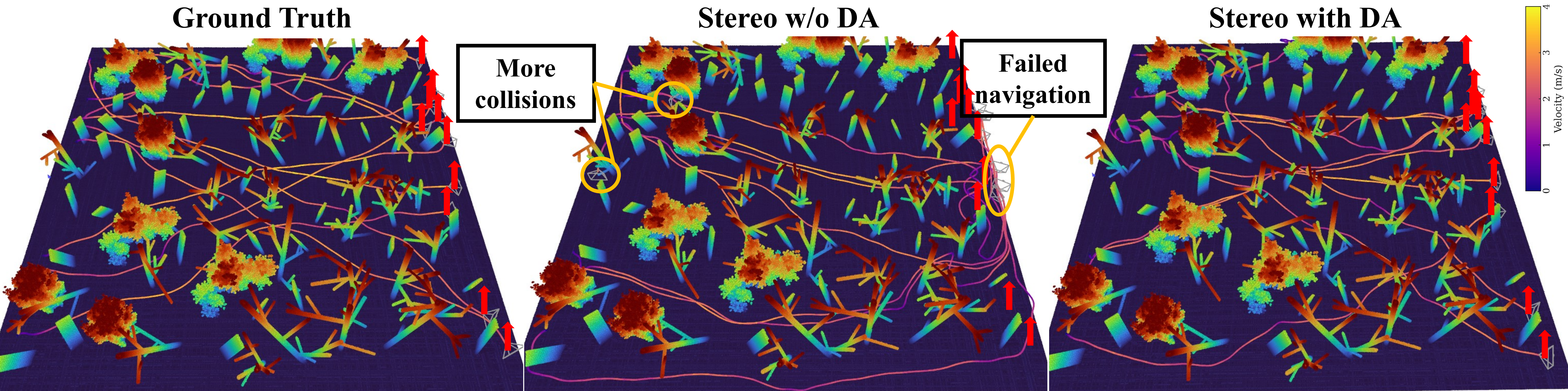}}
    \caption{Policy evaluation in IsaacGym with different depth inputs. The left column shows evaluation with ground truth depth images, the middle column shows the policy evaluated with stereo depth images without encoder refinement, and the right column shows the policy evaluated with stereo depth images after domain adaptation.}
    \label{fig: rerun}
    \vspace{-4mm}
\end{figure*}

As shown in Figure~\ref{fig: rerun}, the policy trained on ground truth depth performs well when evaluated on the same type of input (left), but fails to generalize to stereo depth images without encoder refinement (middle), resulting in frequent collisions and missed targets. After applying domain adaptation (right), the latent space is better aligned, enabling successful navigation.

We also trained a stereo baseline policy using stereo depth images from IsaacGym. Due to the increased computational complexity of training with stereo depth images, this policy required approximately 50 hours to converge, whereas the policy trained with ground truth depth images completed training in 18 hours. All training was conducted on an RTX 4090 GPU.

The evaluation results are summarized in Table~\ref{tab:results}. The \textit{GT} column represents the upper bound performance, where evaluation is conducted using ground truth depth images. The \textit{Stereo Baseline} column shows the performance of the stereo baseline policy, which is lower than the GT upper bound due to the challenges of training with stereo depth images. The \textit{Stereo} column presents the evaluation results for a policy trained on ground truth depth images but evaluated directly with stereo depth images without any refinement, resulting in a significant performance drop.
The \textit{Refinement w/o DA} column shows results where the encoder is refined solely using a reconstruction loss, without domain adaptation. In contrast, the \textit{Domain Adaptation} column represents a policy trained on ground truth depth images but evaluated with stereo depth images after applying domain adaptation, demonstrating improved generalization.

All the above policies use the VAE latent space as input. The highest success rate is achieved by the \textit{LSTM \& DA} configuration, which incorporates temporal memory via an LSTM and refines the encoder but not the policy using both reconstruction and domain adaptation losses.

\begin{table}[t]
    \setlength{\tabcolsep}{2.5pt} 
    \small 
    \centering
    \caption{Evaluations across different latent space configurations.}
    \label{tab:results}
    \begin{tabular}{l c | c c c c c}
        \toprule
        & GT & \makecell{Stereo \\ Baseline} & Stereo & \makecell{Refinement \\ w/o DA} & \makecell{Domain \\ Adaptation} & \makecell{LSTM \\ \& DA} \\
        \midrule
        \makecell{Success \\ Rate (\%)} & 92.5 & 78.8 & 40.0 & 63.8 & 77.5 & \textbf{81.2} \\
        \bottomrule
    \end{tabular}
    \vspace{-4mm}
\end{table}

\subsection{Zero-shot Policy Transfer in AvoidBench}

\begin{figure}[tpb]
    \centering
    {\includegraphics[width=3.4in]{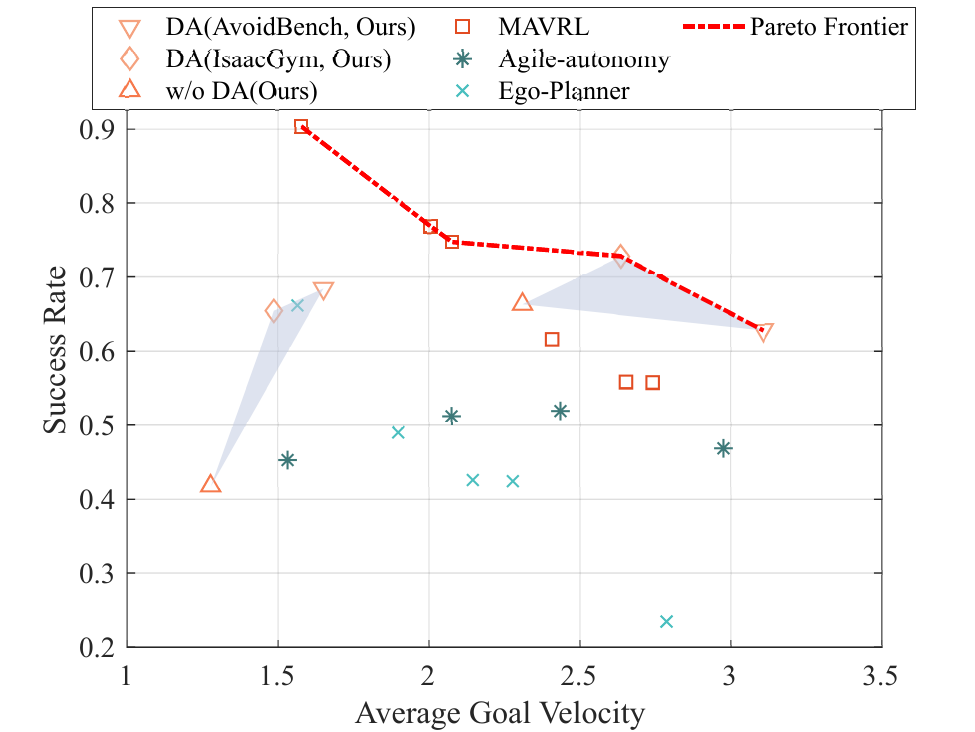}}
    \caption{Pareto front of the proposed method compared to existing approaches.}
    \label{fig: pareto_front}
\vspace{-6mm}
\end{figure}

To evaluate the overall performance of our obstacle avoidance approach, we benchmarked the proposed method using AvoidBench. Two policies with different maximum speed limits were trained by fine-tuning the progress reward hyperparameters in Equation~\ref{eq:progress_reward}. These policies were then evaluated using stereo depth images as input. As shown in Figure~\ref{fig: pareto_front}, the x-axis denotes the average goal velocity~\cite{10161097}, while the y-axis indicates the corresponding success rate. Each marker in the plot represents aggregated results over 1260 trials (30 trials per map across 42 maps).

Each policy is evaluated using three encoder configurations: an upward-pointing triangle denotes the encoder trained solely on ground truth depth images from IsaacGym without domain adaptation; a diamond-shaped marker represents the encoder refined with stereo depth images from IsaacGym; and a downward-pointing triangle indicates the encoder refined with stereo depth images from AvoidBench. Although the policies are trained in IsaacGym using ground truth depth images, the encoder refinement is conducted separately using stereo depth data from different sources. The two blue-shaded triangles in the figure correspond to the two policies with different speed profiles, with each triangle's vertices representing the three encoder configurations.


Square markers represent MAVRL~\cite{10816139}, which trains both the encoder and policy directly in AvoidBench using stereo depth. Star and cross markers indicate Agile-Autonomy~\cite{Loquercio2021Science} and Ego-Planner~\cite{zhou2020ego}, respectively. MAVRL performs best in low-speed scenarios (average velocity \(<2.1~m/s\)), likely due to full training in AvoidBench. In contrast, our method dominates the high-speed regime, benefiting from a reference generator that ensures consistent trajectory tracking and stable flight. Despite being trained entirely in IsaacGym, our policy generalizes effectively via domain adaptation—even when refined with stereo depth from different sources.

Focusing on the triangle vertices, we observe that using stereo depth from AvoidBench for encoder refinement consistently yields the highest goal velocity, closely matching the behavior seen during training. At low speeds, refinement with AvoidBench stereo depth leads to the highest success rate, whereas at high speeds, refinement with IsaacGym stereo depth yields superior performance. These results highlight the effectiveness of our domain adaptation method in aligning latent spaces across domains, enabling robust zero-shot policy transfer across simulation and real-world environments.

\subsection{Zero-shot Policy Transfer in Real-world Environments}

\begin{figure*}[tpb]
    \centering
    \begin{subfigure}[t]{0.78\textwidth}
        \centering
        \includegraphics[width=\linewidth]{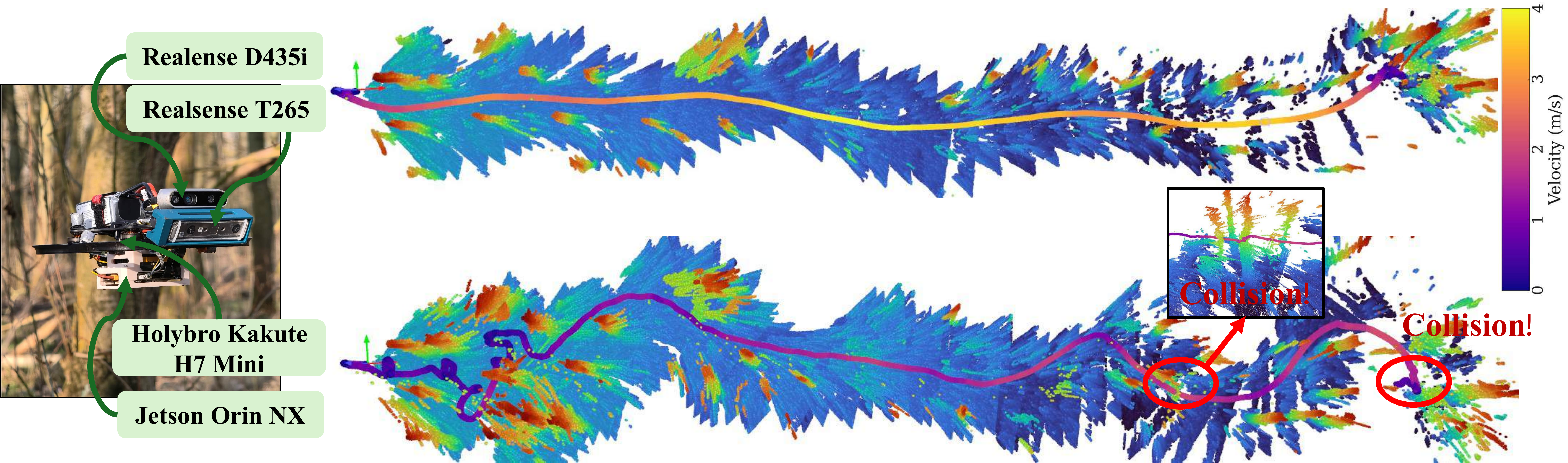}
        \caption{}
        \label{fig: real_world}
    \end{subfigure}
    \hfill
    \begin{subfigure}[t]{0.2\textwidth}
        \centering
        \includegraphics[width=\linewidth]{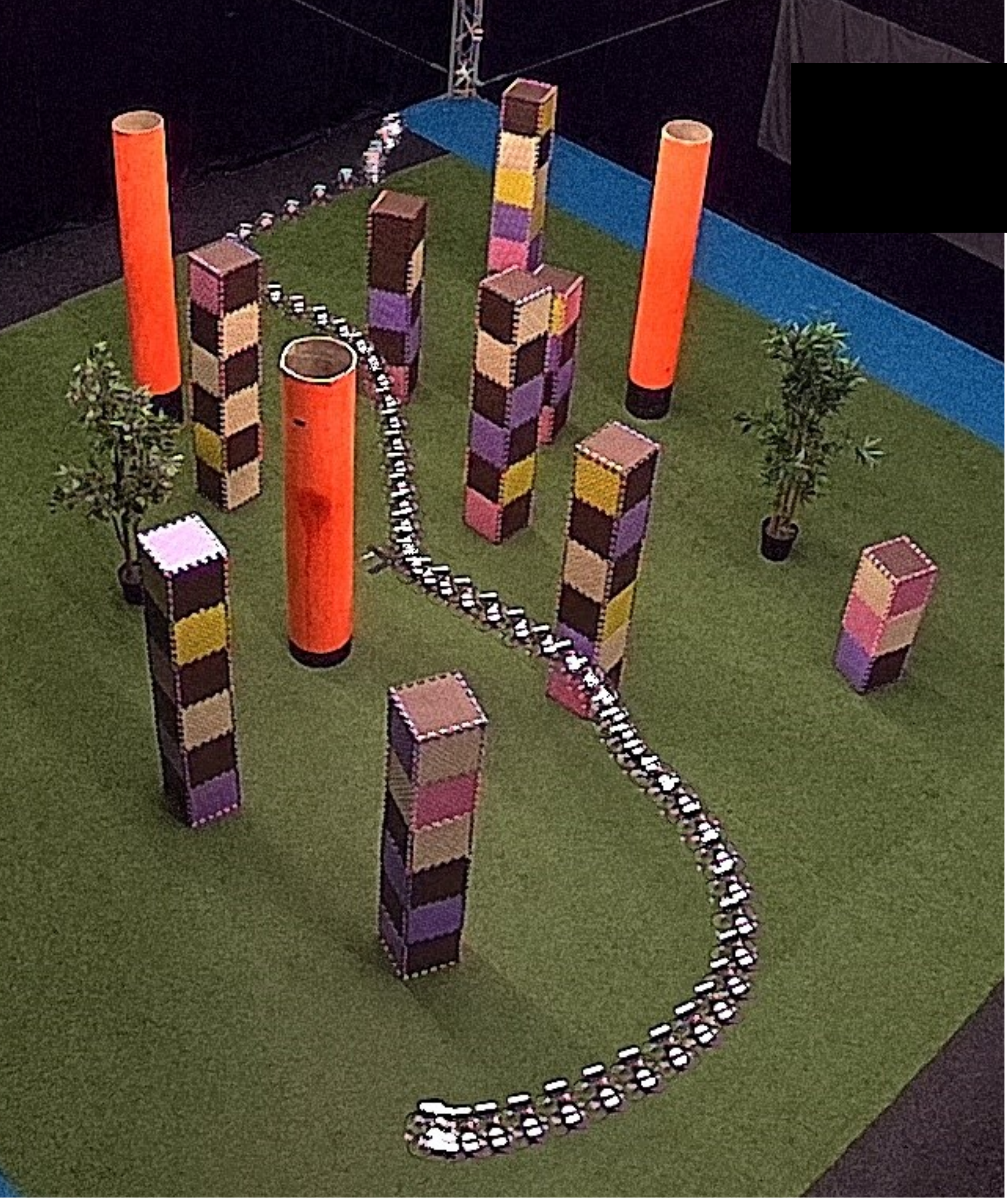}
        \caption{}
        \label{fig: cyberzoo}
    \end{subfigure}
    \caption{Real-world experiments conducted in both outdoor (a) and indoor (b) environments demonstrate the effectiveness and generalization of our domain-adapted depth transfer method.}
    \label{fig:real_world_combined}
    \vspace{-4mm}
\end{figure*}

To validate our depth transfer method in real-world scenarios, we deploy the trained policy on a small drone with a \SI{150}{mm} wheelbase, equipped with an Intel RealSense D435i for depth sensing and a T265 for state estimation. The low-level controller runs on a Holybro Kakute H7 Mini with Betaflight, while the high-level policy runs on a Jetson Orin NX. We adopt Agilicious~\cite{foehn2022agilicious} as the high-level control framework, and use the same reference generator as in simulation, with Model Predictive Control (MPC) for accurate tracking.

Figure \ref{fig: real_world} illustrates the real-world flight trajectories of the drone using two different encoders: one trained on IsaacGym ground truth depth images (bottom trajectory) and another refined with domain adaptation (top trajectory). The results demonstrate that domain adaptation enables direct transfer of the policy trained in IsaacGym to real-world environments, allowing the drone to successfully avoid obstacles and reach the target. In contrast, when using the unrefined encoder, the drone exhibited erratic lateral movements in dense obstacle regions and collided twice, while also maintaining a lower overall speed compared to the domain-adapted encoder.

We also test the same policy in a challenging indoor environment (Figure~\ref{fig: cyberzoo}) using the encoder refined with outdoor stereo depth. The drone navigates narrow corridors and avoids obstacles, demonstrating strong generalization of our method to diverse real-world settings.


\section{CONCLUSION}
Our work introduces a novel depth transfer method for training reinforcement learning policies in simulation and deploying them in real-world environments. By leveraging a VAE with an LSTM, we extract latent representations from depth images, allowing the policy to capture temporal dependencies and generalize across diverse environments. To enhance policy transferability, we propose a domain adaptation technique that aligns the latent spaces of simulated and real-world depth images. Experimental results demonstrate that domain adaptation significantly improves policy performance, enabling zero-shot policy transfer across different environments. 


\bibliographystyle{IEEEtran}
\bibliography{IEEEexample}

\end{document}